\documentclass[10pt,twocolumn,letterpaper]{article}
\usepackage{amsmath}
\usepackage{amsfonts}
\usepackage{amssymb}
\usepackage{algorithm}
\usepackage{algorithmic}
\usepackage{multirow}
\usepackage{pifont}
\usepackage{colortbl}
\usepackage[pagenumbers]{iccv} % To force page numbers, e.g. for an arXiv version

\definecolor{iccvblue}{rgb}{0.21,0.49,0.74}
\usepackage[pagebackref,breaklinks,colorlinks,allcolors=iccvblue]{hyperref}
\title{UniV2D: Bridging Visual Restoration and Semantic Perception for Underwater Salient Object Detection}

\author{
    Laibin Chang$^{1}$, Shaodong Wang$^{1}$, Yunke Wang$^{2}$, Xu Zhang$^{1}$, Kui Jiang$^{3}$, Chang Xu$^{2}$, Bo Du$^{1}$\\
    $^{1}$ School of Computer Science, Wuhan University \\
    $^{2}$ School of Computer Science, The University of Sydney \\
    $^{3}$ School of Computer Science and Technology, Harbin Institute of Technology\\
}
\begin{document}
\maketitle
\begin{abstract}
  Underwater salient object detection (USOD) plays a vital role in marine vision tasks but remains fundamentally challenging due to severe visual degradation, such as selective absorption and medium scattering. Conventional pipelines typically adopt a sequential "enhance-then-detect" paradigm. However, isolating low-level visual restoration from high-level semantic perception often leads to semantic inconsistency, where the restored images may not be optimal for detection and can even introduce task-irrelevant noise. To break this sequential bottleneck, we propose UniV2D, a Unified Vision-to-Detection Network that jointly optimizes visual restoration and salient object detection within a mutually beneficial framework. Unlike traditional methods that rely on disjointed pipelines or rigid physical priors, UniV2D introduces a semantic-driven learning paradigm: high-level saliency semantics actively guide the restoration process, while the restored visual cues reciprocally enhance saliency perception. Specifically, UniV2D features a hierarchical dual-branch architecture. It first employs a self-calibrated decoder to predict initial saliency masks alongside a mask-aware restoration module to reconstruct image content. Subsequently, a saliency-guided refinement module equipped with cross-level modulation is utilized to align structural fidelity with semantic consistency. Extensive experiments across multiple benchmarks demonstrate that UniV2D significantly outperforms state-of-the-art methods in both quantitative and qualitative evaluations, establishing a new standard for joint underwater perception.
\end{abstract}
\section{Introduction}\label{Introduction}
\begin{figure}[!tp]
	\setlength{\abovecaptionskip}{0.1cm}
	\setlength{\belowcaptionskip}{0.0cm}
	\centering
	\includegraphics[width=0.466\textwidth]{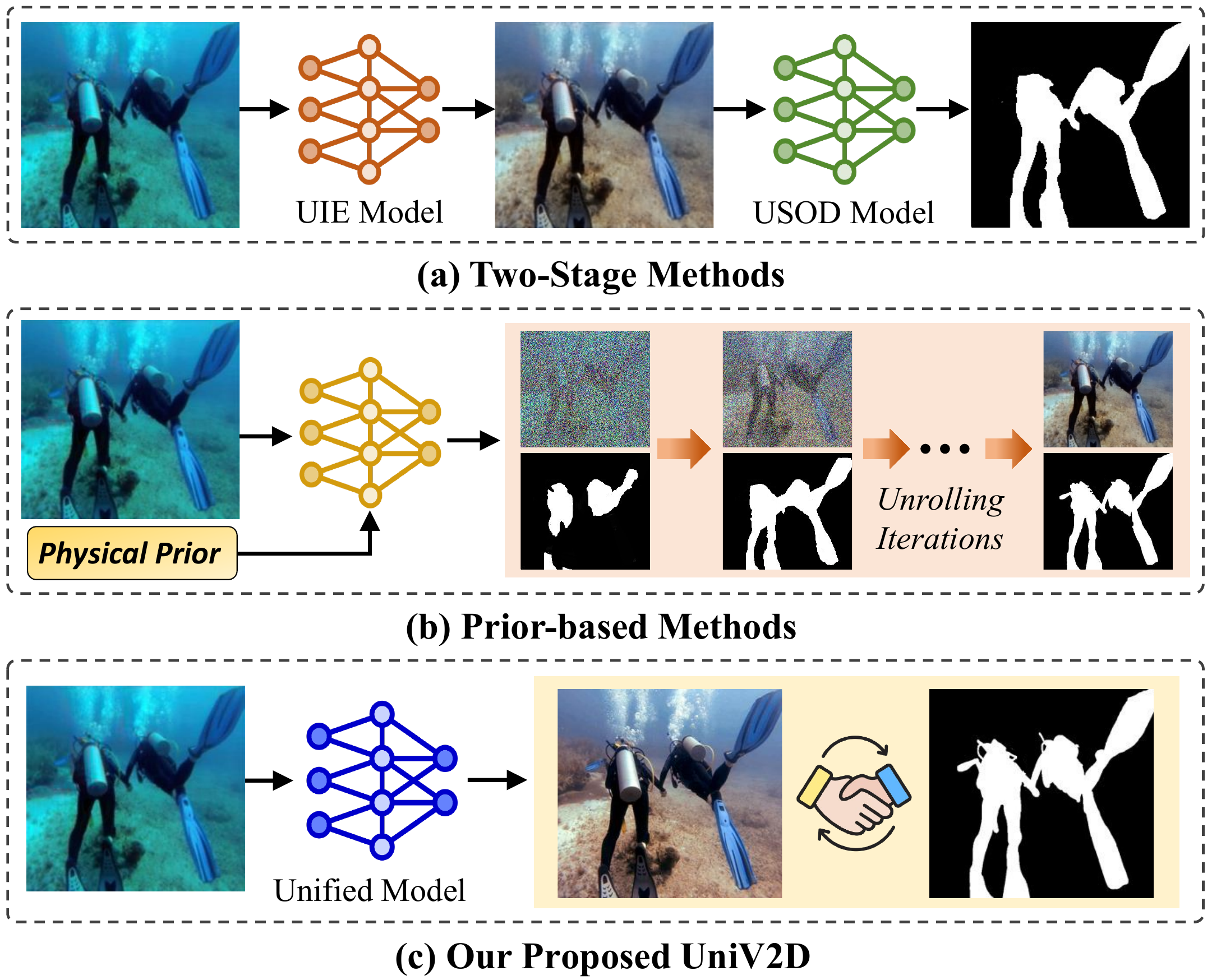}
	\caption
	{Comparison of different paradigms for joint Underwater Image Enhancement (UIE) and Salient Object Detection (USOD). (a) Two-stage methods decouple UIE and USOD into independent networks, resulting in limited cross-task synergy. (b) Prior-based methods (\textit{e.g.}, WaterDiffusion) require auxiliary physical priors and computationally expensive iterative sampling. (c) Our UniV2D establishes a unified framework for simultaneous UIE and USOD, achieving explicit task interaction and efficient one-shot inference.}
	\label{0-1The_First_Image}
\end{figure}

Salient Object Detection (SOD) \cite{Wang2023Pixels, Gao2024Multi} aims to identify and segment the most visually prominent regions within an image, typically corresponding to foreground objects that attract human attention \cite{Hao2025Simple, Zhu2025Dc, Hu2024Cross, Tang2024Divide, Lin2026SamDAQ, Cong2025Trnet, Cong2023Point, Cong2025Breaking}. Although considerable progress has been achieved in terrestrial SOD tasks, largely facilitated by large-scale annotated datasets and advances in deep learning architectures, the performance of these models deteriorates markedly when applied to underwater environments \cite{Ma2025Stamf, Zhang2021Cross}. This decline primarily stems from the inherent challenges of underwater imaging, including light attenuation, color distortion, and low contrast, which collectively lead to the loss of texture and fine details \cite{Li2025Fscdiff, Li2025Efficient, Chang2023ISPRS, Perceive-IR}. Such degradations severely weaken visual saliency cues, making it increasingly difficult for conventional models to accurately detect and localize salient objects in complex underwater scenes.

Underwater Salient Object Detection (USOD) has garnered increasing attention due to its strategic importance in various vision-based marine exploration applications \cite{Hong2023Vision, Zheng2024Marineinst, Lian2024Diving, Peng2024Blurriness, Zhou2026Prompt, Lian2023Watermask}. Despite its potential, USOD faces several inherent challenges: 1) Poor visibility in underwater imagery impedes the accurate localization of salient regions \cite{Hong2025Usis16k, Li2025Uwsam, Chang2026Color}; 2) Underwater devices with limited computational resources are not suitable for deploying high-complexity models \cite{Wu2024Effiseanet, Liu2024Auto, Wu2025High, Chang2025Marine}. To mitigate these issues, existing USOD methods \cite{Jin2025SPDE, Yang2023Saliency, Liu2025Hdanet, Yuan2025If, Zhu2025Saliency} commonly incorporate handcrafted underwater imaging priors (such as transmission or depth maps) to compensate for visual artifacts. However, such heavy reliance on physical priors limits model generalization, as these priors are frequently unavailable or unreliable in diverse real-world conditions. Furthermore, while several sophisticated architectures \cite{Zha2025Heterogeneous, Deng2023RMFormer, Hong2025USOD10k} have pushed the boundaries of detection performance, they inevitably introduce excessive parameter counts and computational overhead, which significantly restricts their practical deployment on resource-constrained underwater devices.

To alleviate the interference caused by underwater degradation, a straightforward solution is to adopt a sequential "enhance-then-detect" pipeline, where a dedicated Underwater Image Enhancement (UIE) model pre-processes the image before salient object detection \cite{Liu2022Twin, Chen2020Perceptual}, as illustrated in Fig. \ref{0-1The_First_Image}(a). However, most UIE methods prioritize global perceptual quality and may inadvertently overlook or even distort local saliency cues \cite{Zhang2025Hupe}. This two-stage paradigm fundamentally decouples low-level visual restoration from high-level semantic perception, lacking an effective mechanism for cross-task collaborative representation learning. While UIE aims to restore visual clarity and USOD requires structural fidelity for precise localization, these two objectives are often not perfectly aligned in disjointed frameworks. Recently, diffusion-based approaches like WaterDiffusion \cite{Chang2025Waterdiffusion} have attempted to bridge this gap by jointly optimizing restoration and detection through iterative unrolling and physical imaging priors, as shown in Fig. \ref{0-1The_First_Image}(b). Although these methods achieve impressive results, their heavy reliance on handcrafted imaging assumptions and the high computational cost of multiple optimization iterations limit their practical efficiency. These limitations motivate us to develop a more efficient collaborative framework that seamlessly integrates visual restoration and salient object detection within a unified architecture.

%Although such designs have achieved promising joint-task performance, it rely on handcrafted underwater imaging assumptions and entail substantial computational overhead, thereby limiting its applicability in real-world underwater scenarios. These considerations motivate the design of a collaborative and lightweight framework that can effectively integrate visual restoration and salient object detection within a unified architecture.

Motivated by the above analysis, we propose UniV2D, a novel and unified framework designed to jointly optimize underwater image restoration and salient object detection in a mutually reinforcing manner. Diverging from the conventional sequential pipelines or those tethered to rigid physical priors, UniV2D introduces a semantic-driven dual-branch architecture that establishes explicit task interaction from the initial decoding stages. Specifically, the Self-Calibrated Saliency Masking (SCSM) module progressively generates a saliency map by incorporating spatial reweighting with token-guided feature calibration. The Mask-Aware Content Restoration (MACR) module leverages this predicted saliency as structural guidance to reconstruct clear images with enhanced content consistency. These preliminary outputs are further refined through a saliency-guided cross-level refinement procedure, which employs Cross-Level Feature Modulation (CLFM) to facilitate bidirectional feature interaction across multiple scales. Owing to its streamlined design and efficient learning paradigm, UniV2D achieves robust performance on both tasks while maintaining a significantly low computational overhead.

Our key contributions are summarized as follows:
\begin{itemize}
	\item We propose UniV2D, a unified and lightweight framework that jointly optimizes underwater image restoration and salient object detection without relying on handcrafted physical priors or iterative optimization.
	
	\item We design a coarse-to-refinement architecture featuring SCSM, MACR, and CLFM modules, which effectively bridges the gap between low-level visual restoration and high-level semantic perception via bidirectional feature interaction.
	
	\item Extensive experiments demonstrate that UniV2D outperforms existing state-of-the-art UIE and USOD solutions in both qualitative and quantitative outcomes.
\end{itemize}

\section{Related Work}
\label{Related Work}
%Despite their success, these methods tend to adopt invariant empirical parameters whose poor generalisability cannot cope with complex and changeable underwater scenes.
\textbf{Underwater Image Enhancement}. UIE is a practical yet challenging task within the field of visual restoration, which mainly includes the physics-based and deep learning-based methods \cite{Wang2024Inspiration, Zhang2025Uniuir, Fan2025Llava, Liu2025Toward, Song2023Dual_model, UniUIR}. These physics-based methods \cite{Chen2025Fusion} aim to improve visual perception by directly manipulating image pixel values with well-designed techniques, regardless of the underwater degradation mechanism, including multi-scale fusion \cite{Song2022ERH, Zhang2024Underwater}, Retinex-based \cite{Zhuang2022Retinex, Zhou2024Pixel}, and histogram equalization \cite{Huang2018RGHS}. Deep learning-based UIE methods \cite{Cao2025Erd} concentrate on enhancing degraded images by autonomously learning non-linear restoration mappings from paired underwater image datasets \cite{ClearAIR, UIC_zhu}. Recently, diffusion-based methods \cite{Bi2025Seadiff, Deng2025Frequency} have been proposed for underwater image restoration, such as UW-DDPM \cite{Lu2023UW-DDPM}, DiffUIE \cite{Qing2024Diffuie}, WF-Diff \cite{Zhao2024Wavelet}, and DCGF \cite{Zhang2025Dcgf}. Additionally, several domain adversarial learning networks have been developed to improve the adaptability of UIE models, including TUDA \cite{Wang2023TUDA}, Semi-Net \cite{Huang2023Semi-UIR}, UICoE-Net \cite{Qi2022JointLearning}, \textit{etc}. While these methods demonstrate a certain level of domain adaptation, they do not explicitly prioritize the enhancement of salient regions. As a result, they remain inadequate for supporting downstream salient object detection tasks.

%Inspired by the diffusion model, Lu \emph{et al.} \cite{Lu2023UW-DDPM} proposed a UIE method called UW-DDPM, which utilizes two U-Net networks to perform denoising and distribution transforms. Then, Tang \emph{et al.} \cite{Tang2023DM-based} introduced a transformer-based diffusion network for UIE. Although the images restored by the diffusion model are superior to the GAN-based generative methods, they are only conditioned on degraded images in diffusion, without caring about saliency generation.

%Inspired by the revised underwater imaging models, several deep learning methods \cite{Wang2017Imaging_model, Chen2020Perceptualmodel, Song2023Dual_model} have been proposed for reconstructing clear underwater images by estimating the ambient-light and direct-transmission parameters, which are superior to the prior estimation based on a single image.
\begin{figure*}[!htp]
	\setlength{\abovecaptionskip}{0.1cm}
	\setlength{\belowcaptionskip}{0.0cm}
	\centering
	\includegraphics[width=1.0\textwidth]{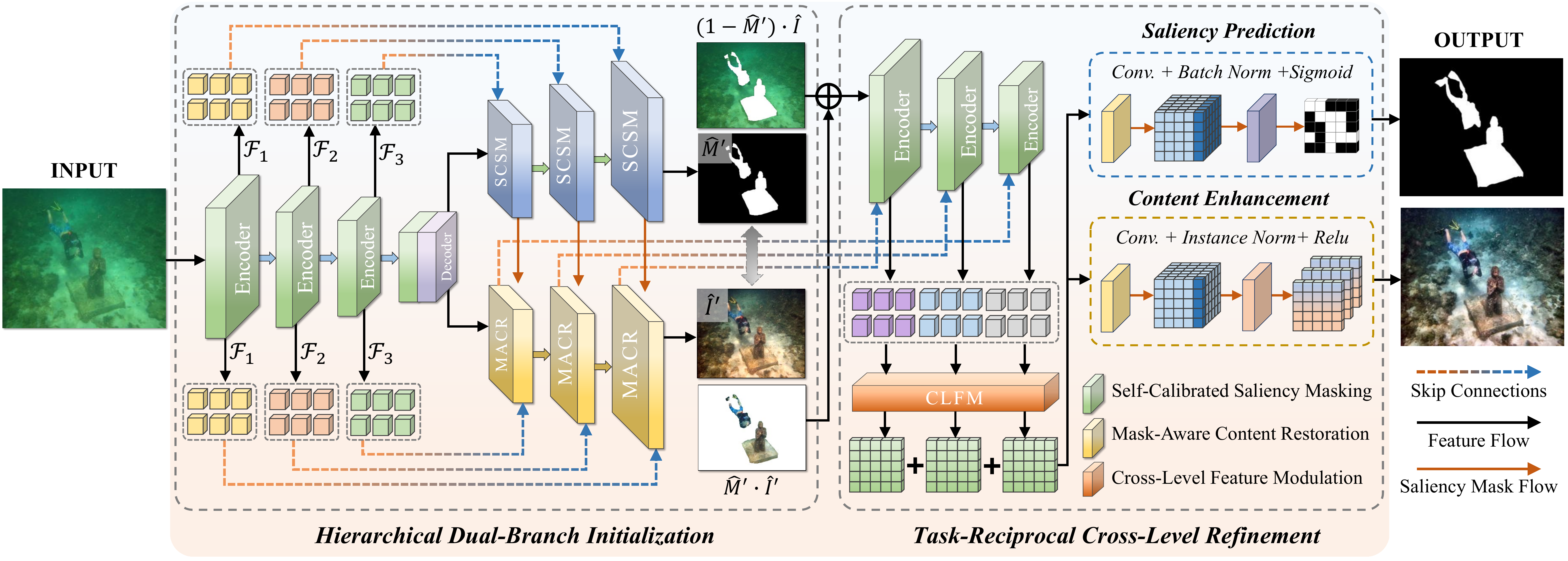}
	\caption
	{Overview of the proposed UniV2D. It features a semantic-driven dual-branch design, consisting of Self-Calibrated Saliency Masking (SCSM) and Mask-Aware Content Restoration (MACR) modules for initial task estimation, followed by the Cross-Level Feature Modulation (CLFM) module for joint task-reciprocal refinement of both tasks.}
	\label{1-1The_Overall_Framework}
\end{figure*}

\noindent\textbf{Underwater Salient Object Detection}. Early feature-based USOD methods \cite{Kumar2019Saliency, Kanwal2024Unveiling} attempt to encode low-level image features (\textit{e.g.}, color, texture, and contour) into descriptors and then infer visual saliency by quantifying their global relative sharpness \cite{He2023Multispectral, He2023Object}. Jian \emph{et al.} \cite{Jian2018Integrating} proposed a framework that combines the quaternionic distance-based Weber descriptor, pattern distinctness, and local contrast to highlight salient objects and suppress background regions. Instead of linearly stacking multiple convolutional layers in the network, several deep learning-based USOD methods \cite{Zha2025Heterogeneous, Hong2025USOD10k, Zhou2026Turbidity} have incorporated visual transformers with wider receptive fields into their deep architectures. This approach helps alleviate the computational burden imposed by convolution and improves the salient object detection performance to some extent \cite{Chen2024Bauodnet, Cong2025Uis}. However, these methods do not consider cross-level feature interactions when designing the network, which is not reliable for capturing saliency cues by improving the encoder architecture.

\section{Methodology}\label{Methodology}
%\subsection{Overall Framework}
The overall architecture of UniV2D is illustrated in Fig. \ref{1-1The_Overall_Framework}. Diverging from conventional two-stage paradigms, UniV2D adopts a unified dual-branch design to jointly optimize underwater image restoration and salient object detection within a single network. Built upon a U-Net-like backbone \cite{Ronneberger2015U_net}, the framework comprises a shared hierarchical encoder and two task-specific decoder branches, enabling efficient feature reuse while mitigating task decoupling issues. The encoder extracts multi-scale features that capture both fine-grained structural details and high-level semantic context, providing a foundation for subsequent collaborative decoding.

The core of our framework lies in a two-stage optimization process: hierarchical dual-branch initialization and task-reciprocal cross-level refinement. During initialization, the saliency branch incorporates the Self-Calibrated Saliency Masking (SCSM) module to progressively localize salient regions. Simultaneously, the restoration branch employs the Mask-Aware Content Restoration (MACR) module, which leverages the predicted saliency maps as structural guidance to reconstruct clear and content-consistent images. Following this, the preliminary outputs are further enhanced through Cross-Level Feature Modulation (CLFM) modules. These modules facilitate bidirectional feature interaction across multiple scales and tasks, effectively reinforcing semantic alignment and structural fidelity. Finally, UniV2D is optimized via a joint loss function that balances saliency detection accuracy and visual restoration quality, ensuring both tasks benefit from shared supervision and mutual reinforcement.

%In the hierarchical dual-branch initialization, the saliency branch incorporates the Self-Calibrated Saliency Masking (SCSM) module to produce refined saliency estimations, while the restoration branch employs the Mask-Aware Content Restoration (MACR) module that leverages the predicted saliency maps as structural priors to enhance texture reconstruction. Both branches consist of three decoder levels, where the saliency branch additionally benefits from side-output supervision to stabilize mask prediction. Then, the coarse predictions from the two branches are refined through the Cross-Level Feature Modulation (CLFM) module, which propagates complementary cues across scales and tasks, thereby improving semantic alignment and structural fidelity. Finally, the network is optimized with joint loss functions that balance saliency detection accuracy and visual restoration quality, ensuring both tasks benefit from shared supervision and collaborative learning.

\subsection{Dual-Branch Initialization via Hierarchical Saliency Decoding}
As the foundational stage of UniV2D, the Dual-Branch Initialization aims to jointly generate a coarse saliency mask and a restored image through a hierarchical decoding process. Diverging from conventional two-stage pipelines that handle saliency detection and image restoration in isolation \cite{Liu2022Twin, Zhou2025Spatial}, our design establishes explicit task interaction at the early decoding stage, allowing the two objectives to mutually guide feature reconstruction. To this end, the decoder is organized into two parallel branches: the SCSM module produces an initial saliency prior, while the MACR module leverages these estimated cues to recover content with structural fidelity. This hierarchical architecture progressively integrates multi-scale encoder features, enabling a coarse-to-fine estimation of both tasks while maintaining intrinsic consistency.
\begin{figure}[!htp]
	\setlength{\abovecaptionskip}{0.1cm}
	\setlength{\belowcaptionskip}{0.0cm}
	\centering
	\includegraphics[width=0.466\textwidth]{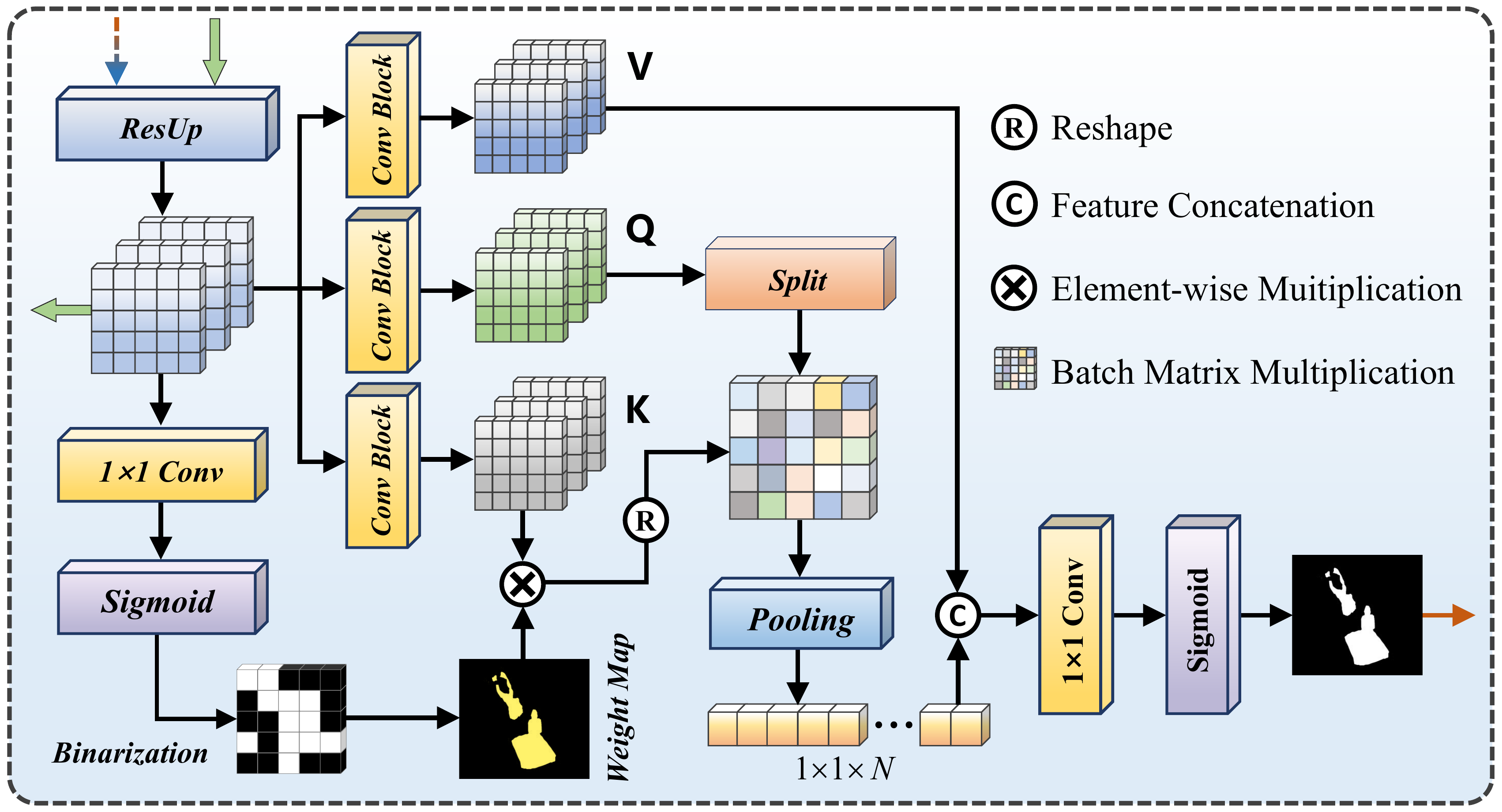}
	\caption
	{Architecture of the SCSM module. It employs spatial reweighting and token-guided calibration to produce an initial saliency mask from hierarchical features.}
	\label{2-1Self-Calibrated_Saliency_Masking}
\end{figure}

\textbf{Self-Calibrated Saliency Masking.}
The SCSM module is designed to generate an adaptive saliency prior that emphasizes foreground regions while suppressing background interference. As illustrated in Fig. \ref{2-1Self-Calibrated_Saliency_Masking}, the decoder feature $\mathcal{S}_i$ is first fused with its corresponding skip-connected encoder feature $\mathcal{F}_i$ via channel-wise concatenation. This is followed by a residual upsampling $\text{ResUp}(\cdot)$ to align the spatial resolution with the preceding layer:
\begin{equation}\label{eq:x_in}
\hat{X}_\textit{in} = \mathrm{ResUp}\left(\mathrm{Cat} \left(\mathcal{S}_i,\mathcal{F}_i \right)\right).
\end{equation}

Based on the fused feature $\hat{X}_\textit{in}$, we estimate a coarse saliency weighting map $\mathcal{M}$ using a $1{\times}1$ convolution to aggregate channel responses, followed by a Sigmoid activation that normalizes the output to the range $[0,1]$. The computation is defined as:
\begin{equation}\label{eq:coarse}
\mathcal{M}_\textit{k} = \sigma \left( \text{Conv}_{1 \times 1}\left(\hat{X}_\textit{in}\right) \right),
\end{equation}
where $\sigma(\cdot)$ denotes the Sigmoid function. The resulting $\mathcal{M}_\textit{k}$ serves as a task-adaptive prior, effectively enhancing salient responses while filtering out pervasive underwater background noise.

Although $\mathcal{M}_\textit{k}$ may still contain background noise or incomplete foreground responses, it serves as a task-adaptive prior that highlights informative regions while suppressing irrelevant areas. Guided by this prior, we derive a compact global salient descriptor $\hat{X}_k \in \mathbb{R}^{C}$ through attention-weighted pooling over the feature map $\hat{X}_\textit{in}$, formulated as:
\begin{equation}
\hat{X}_\textit{k} = \frac{1}{\sum_{(i,j)} \mathcal{M}^{(i,j)}_\textit{k}} \sum_{(i,j)} \mathcal{M}^{(i,j)}_\textit{k} \cdot \hat{X}_\textit{in}^{(i,j)},
\end{equation}
where $\mathcal{M}_\textit{k}^{(i,j)}$ serves as an adaptive weight to effectively aggregate discriminative cues from salient regions.

To evaluate the semantic consistency between global salient features and local spatial features, we project both $\hat{X}_\textit{k}$ and $\hat{X}_\textit{in}$ into a shared representation space using two learnable convolutional blocks $\phi_\textit{q}$ and $\phi_\textit{k}$, as defined below:
\begin{equation}
\tilde{X}_\textit{k} = \phi_\textit{k}(\hat{X}_\textit{k}), \quad \tilde{X}_\textit{q} = \phi_{q}(\hat{X}_\textit{in}),
\end{equation}
where $\tilde{X}_\textit{k}$ and $\tilde{X}_\textit{q}$ denote the transformed feature maps within the common embedding space. $\phi_\textit{q}(\cdot)$ and $\phi_\textit{k}(\cdot)$ are $1{\times}1$ convolutional blocks that align the local and global representations into a comparable feature domain.  

We then calculate a saliency affinity map $S\in \mathbb{R}^{1 \times H \times W}$ to quantify the semantic similarity between each pixel and the global salient region. This is achieved by concatenating $\tilde{X}_\textit{k}$ and $\tilde{X}_\textit{q}$ along the channel dimension, followed by a $1{\times}1$ convolution and non-linear activation:
\begin{equation}
S = \sigma\left( \text{Conv}_{1 \times 1} \left( \tanh \left(\text{Cat}(\tilde{X}_\textit{k}, \tilde{X}_\textit{q}) \right) \right) \right),
\end{equation}
where $\tanh(\cdot)$ introduces non-linear feature separation to enhance the response contrast between salient and non-salient regions. The $\sigma(\cdot)$ function further maps the activation into the range $[0, 1]$, yielding a probabilistic saliency confidence map. Higher $S$ values indicate stronger semantic relevance to the globally salient region, thereby promoting structurally consistent saliency refinement.

Following the derivation of the affinity map $S$, we integrate a global contextual cue $\tilde{X}_\textit{v}$, which is obtained by projecting the fused feature $\hat{X}_{in}$ through an additional learnable convolutional block $\phi_v(\cdot)$. The initial saliency prediction $\hat{M}'$ is then generated by concatenating $S$ and $\tilde{X}_v$, followed by a $1{\times}1$ convolution and a Sigmoid activation:
\begin{equation}
\hat{M}^{'}= \sigma\left( \text{Conv}_{1 \times 1}\left(\text{Cat}(S, \tilde{X}_\textit{v})\right) \right).
\end{equation}

This predicted mask $\hat{M}'$ serves as an explicit structural prior for the subsequent restoration branch and is further utilized in the refinement stage to facilitate cross-task interaction. To ensure accurate foreground localization, we supervise this prediction using a binary cross-entropy loss against the ground-truth mask $M_\textit{gt}$. The preliminary saliency loss is defined as:
\begin{equation}\label{binary cross-entropy loss}
\mathcal{L}^{\textit{pre}}_{\textit{mask}}=-\sum_{(i,j)}\left(M_\textit{gt}\log \hat{M}^{'}_{(i,j)} + (1 - M_\textit{gt}) \log (1 - \hat{M}^{'}_{(i,j)}) \right),
\end{equation}
where $\hat{M}^{'}_{(i,j)}$ denotes the predicted saliency at pixel $(i,j)$.

\begin{figure}[!tp]
	\setlength{\abovecaptionskip}{0.1cm}
	\setlength{\belowcaptionskip}{-0.2cm}
	\centering
	\includegraphics[width=0.466\textwidth]{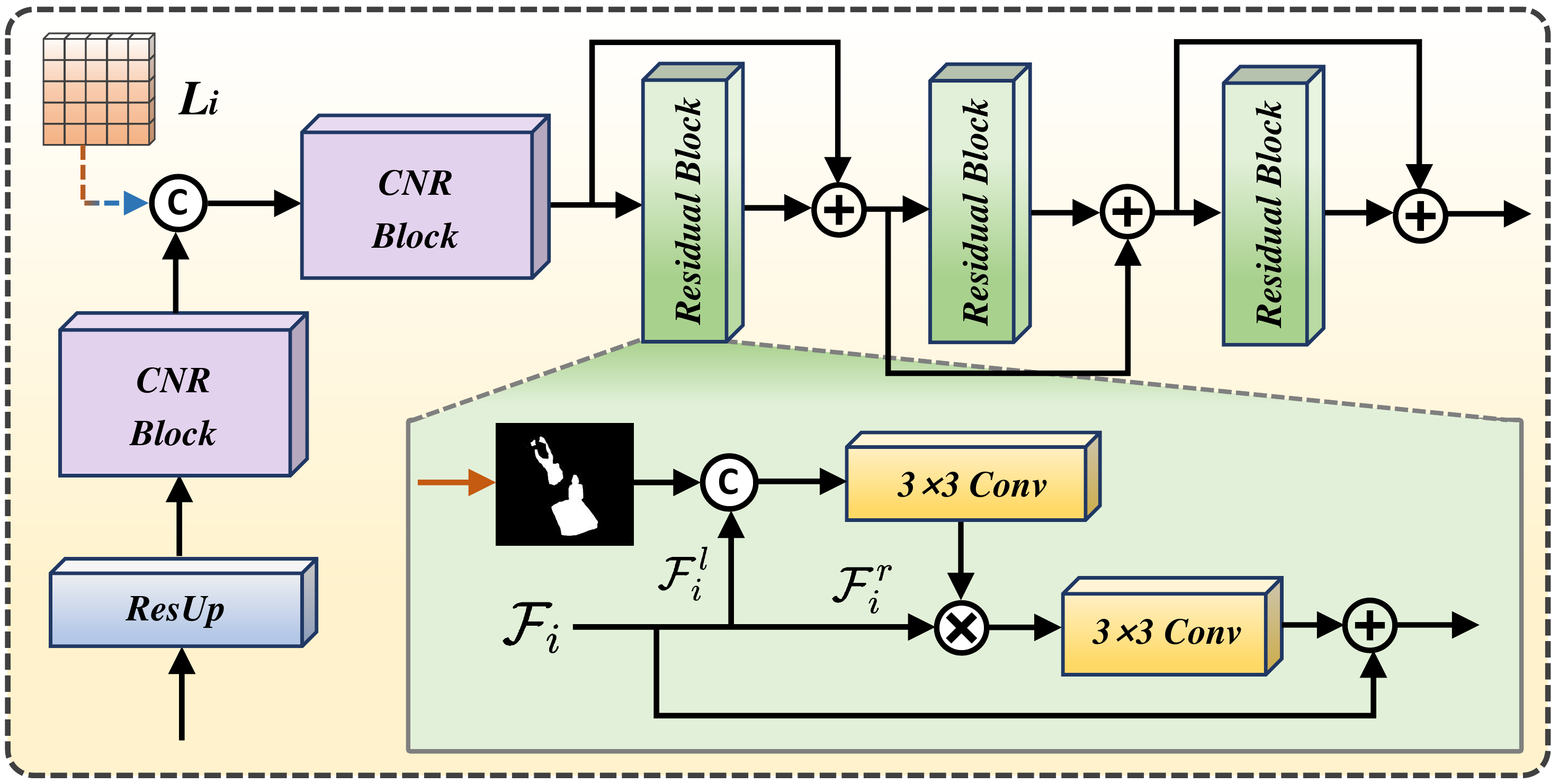}
	\caption
	{Architecture of the MACR module. It utilizes the predicted saliency mask as structural guidance to reconstruct restored images through mask-aware feature modulation.}
	\label{2-2Mask-Aware_Saliency_Enhancement}
	\vspace{-0.2cm}
\end{figure}

\textbf{Mask-Aware Content Restoration.} 
As illustrated in Fig. \ref{2-2Mask-Aware_Saliency_Enhancement}, the MACR module is designed to achieve structure-preserving restoration by leveraging the predicted saliency mask as an explicit spatial prior. Unlike previous multi-task paradigms \cite{Chang2025Waterdiffusion} that rely on simple feature concatenation, MACR adaptively modulates restoration features to emphasize foreground structures while suppressing pervasive underwater background interference.

Given the intermediate encoder feature $\mathcal{F}_i$ from the $i$-th encoder and the propagated feature $\mathcal{G}_{i-1}$ from the previous decoding stage, MACR first aligns the spatial resolution and refines the representation through a residual upsampling and normalization block:
\begin{equation}
\mathcal{G}_i = \text{Cat}\left(\text{CNR}\left(\text{ResUp}(\mathcal{G}_{i-1})\right), \mathcal{F}_i\right),
\end{equation}
where $\text{CNR}(\cdot)$ refers to a sequence of Convolution, Normalization, and ReLU activation.

To incorporate saliency-guided structural information at each level, we concatenate the decoding feature $\mathcal{G}_i$ with the initial saliency mask $\hat{M}'$. We then employ a series of residual blocks to extract a saliency-aware weight map $\mathcal{W}_i$, which serves as a mask-modulated attention mechanism:
\begin{equation}
\mathcal{W}_i = \sigma\left(\text{Conv}_{3\times 3}\left(\text{ReLU}\left(\text{Conv}_{3\times3}\left[\mathcal{G}_i^\textit{l}; \hat{{M}}^{'}\right]\right)\right)\right),
\end{equation}
where $\mathcal{G}_i^\textit{l}$ represents the primary feature partition of $\mathcal{G}_i$, and $\sigma(\cdot)$ denotes the Sigmoid activation. This weight map $\mathcal{W}_i$ effectively encapsulates the spatial importance of different regions, ensuring that the subsequent restoration process is conditioned on the semantic importance of the scene.

The generated weight map $\mathcal{W}_i$ is employed to modulate the remaining feature partition $\mathcal{G}_i^r$ (the right-half channels) through an attention-based residual fusion process. Specifically, $\mathcal{W}_i$ acts as a spatial-aware modulator that is element-wise multiplied with $\mathcal{G}_i^r$ to extract refined structural information. This modulated representation is then processed via a $3 \times 3$ convolution to produce a mask-enhanced residual $\mathcal{R}_i$, calculated as:
%Meanwhile, the generated weight map $\mathcal{W}_k$ also participates in an explicit modeling process through a parallel pathway. Specifically, the $\mathcal{W}_k$ is first projected into a multi-channel attention map, which is then element-wise multiplied with the remaining feature channels (\textit{i.e.}, the right half of the feature map) to extract structural residual information. This residual is subsequently fused with the original feature via a 3×3 convolution to produce a mask-enhanced residual, calculated as:
\begin{equation}
\mathcal{R}_i = \text{Conv}_{3\times3}\left(\mathcal{G}_i^\textit{r} \odot {W}_i\right),
\end{equation}
where $\odot$ denotes element-wise multiplication. By multiplying the feature with the weight map, the network selectively amplifies foreground textures while attenuating background clutter. To preserve the original context while incorporating these saliency-guided updates, the refined feature $\tilde{\mathcal{F}}_i$ is generated through residual aggregation and normalization:
\begin{equation}
\tilde{\mathcal{F}}_i = \text{ReLU}(\text{BN}(\mathcal{R}_i + \mathcal{G}_i)),
\end{equation}
where $\mathrm{BN}(\cdot)$ denotes Batch Normalization. This design ensures that the restoration branch remains sensitive to the semantic boundaries provided by the saliency branch, leading to more coherent and structure-aware underwater image enhancement.

%To stabilize training and gradually inject mask-aware priors, we repeat the residual fusion process over three stacked blocks. Unlike previous multi-task frameworks \cite{Chang2025Waterdiffusion} that directly fuse task features, our MACR module leverages saliency masks as spatial priors to learn structure-aware residuals, achieving better texture fidelity and structural consistency in underwater restoration.
%To stabilize training and progressively inject mask-aware priors, we repeat this residual fusion process across three stacked blocks. Compared to previous multi-task structures \cite{Zhang2025Hupe} that directly fuse features across tasks, our MACR module explicitly incorporates saliency masks as spatial priors and learns structure-specific residuals, yielding superior texture fidelity and structural consistency in underwater restoration.

\subsection{Task-Reciprocal Cross-Level Refinement}
Although the dual-branch initialization yields preliminary results, the restored images may still exhibit residual deficiencies such as blur, artifacts, and color distortion. Simultaneously, the initial saliency mask often suffers from incomplete object responses, which limit its effectiveness as a structural guide. To address these issues, we propose the Task-Reciprocal Cross-Level Refinement stage to achieve global semantic consistency and finer structural alignment through bidirectional task interaction.

As illustrated in Fig. \ref{1-1The_Overall_Framework}, we first integrate the coarse restored image $\hat{I}'$, the initial saliency mask $\hat{M}'$, and the raw input $I$ to form a saliency-aware composite $\tilde{I}_{\textit{pre}}$. This fusion strategy explicitly reinforces content consistency by anchoring the restoration to the predicted salient regions:
%We concatenate the coarse restored image  $\hat{I}^{'}$ with the saliency mask $\hat{M}^{'}$ and feed them into a multi-scale encoder to extract hierarchical features. These features are fused with multi-level decoding features from the initial branch via skip connections. In other words, our approach employs symmetric connections across multiple levels to better integrate structural cues from both image and mask. To further enhance content consistency in salient regions, we introduce a mask-aware reconstruction fusion strategy that combines the restored image $\hat{I}^{'}$, saliency mask $\hat{M}^{'}$, and raw image $\hat{I}$. The fusion is defined as:
\begin{equation}
\tilde{I}_{\textit{pre}} = \hat{I}^{'} \cdot \hat{M}^{'} + (1 - \hat{M}^{'}) \cdot \hat{I},
\end{equation}
where $\tilde{I}{\textit{pre}}$ serves as an auxiliary representation that emphasizes task-relevant regions for the subsequent refinement. This composite is then processed by a dedicated multi-scale encoder to extract deep hierarchical features, which are fused with the decoding features from the initialization stage via symmetric skip connections to preserve multi-level structural cues.
\begin{figure}[!htp]
	\setlength{\abovecaptionskip}{0.1cm}
	\setlength{\belowcaptionskip}{0.1cm}
	\centering
	\includegraphics[width=0.476\textwidth]{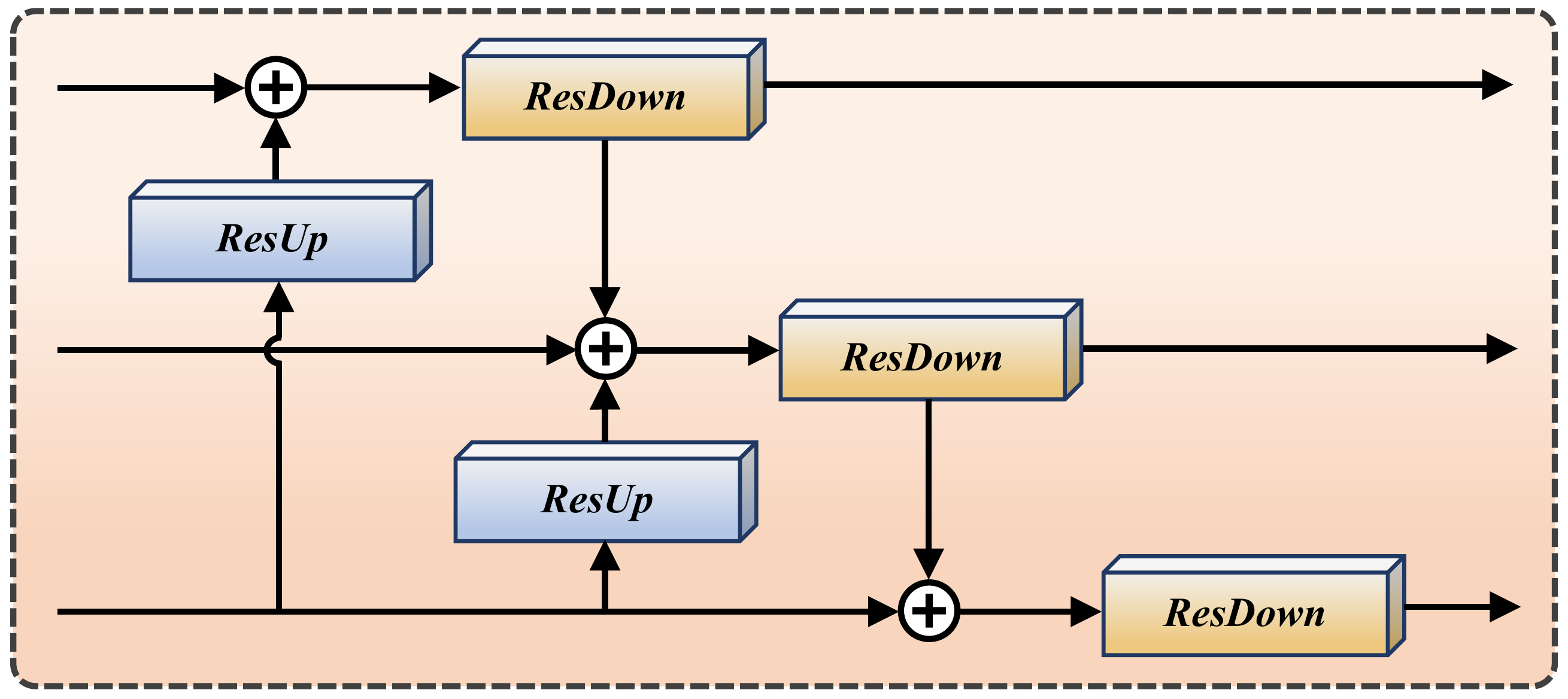}
	\caption
	{Architecture of the CLFM module. It facilitates bidirectional feature interaction between restoration and detection branches across multiple scales.}
	\label{2-3Cross-Level_Feature_Fusion}
    \vspace{-0.2cm}
\end{figure}

\textbf{Cross-Level Feature Modulation.}
To establish effective interaction across these hierarchical representations, we introduce the Cross-Level Feature Modulation (CLFM) module, as shown in Fig. \ref{2-3Cross-Level_Feature_Fusion}. Shallow features capture fine-grained textures, whereas deeper layers encode abstract semantic context. CLFM facilitates efficient cross-scale transfer by integrating task-specific cues through a series of residual-based modulation operations.
\begin{figure*}[!htp]
	\setlength{\abovecaptionskip}{0.05cm}
	\setlength{\belowcaptionskip}{0.0cm}
	\centering
	\includegraphics[width=1.0\textwidth]{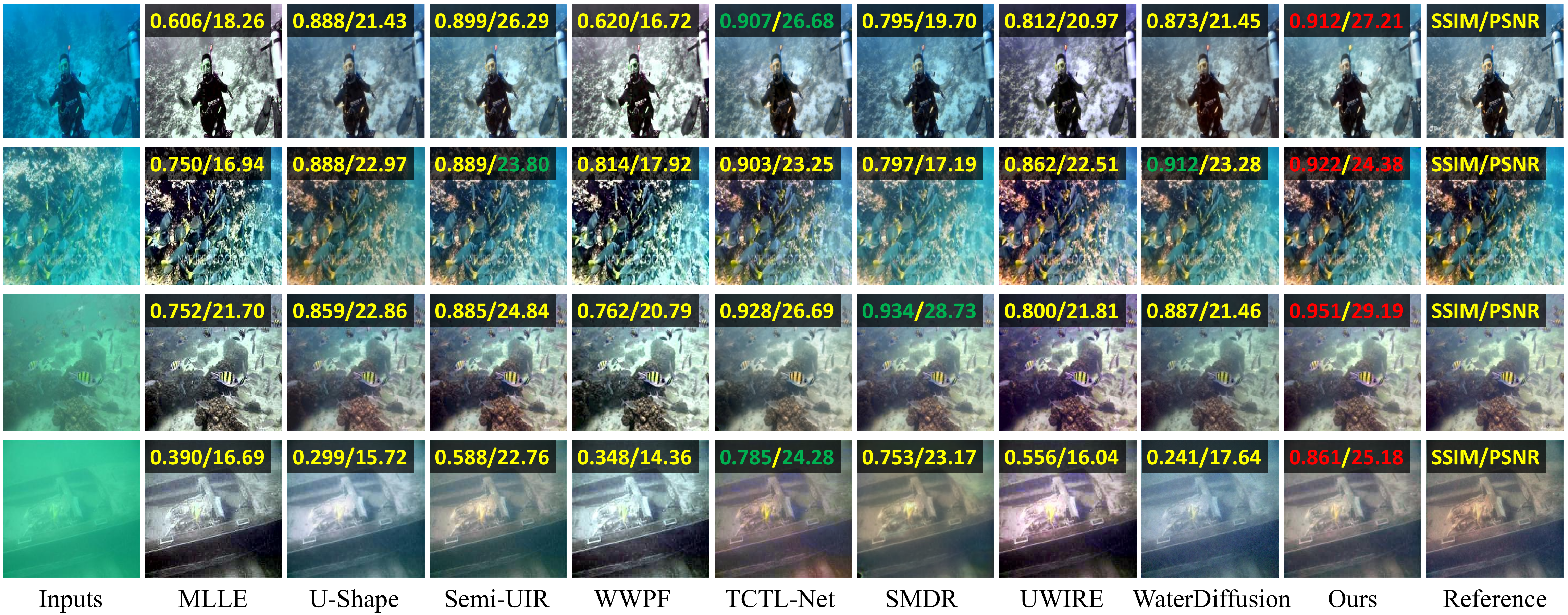}
	\caption	
	{Qualitative comparisons between UniV2D and SOTA methods across diverse underwater degradation types (\textit{e.g.}, greenish, bluish, and turbid conditions). Best and second-best SSIM/PSNR scores are highlighted in \textcolor{red}{red} and \textcolor{green}{green}, respectively.}
	\label{4-1Qualitative_Evaluation_of_Visual_Enhancement}
    \vspace{-0.2cm}
\end{figure*}

The CLFM module utilizes three pairs of residual upsampling ($ResUp$) and downsampling ($ResDown$) paths to align spatial resolutions and adapt channel dimensions across levels. Let $\{\mathcal{F}_{1}, \mathcal{F}_{2}, \mathcal{F}_{3}\}$ denote the encoder features from low to high levels. We progressively modulate these features to propagate semantic and structural information as follows:
\begin{align}
\mathcal{F}^d_1 & = {ResDown}(\mathcal{F}_1 + {ResUp}(\mathcal{F}_3)),\\ 
\mathcal{F}^d_2 & = {ResDown}(\mathcal{F}^d_1 + {ResUp}(\mathcal{F}_3)+ \mathcal{F}_{2}),\\
\mathcal{F}^d_3 & = {ResDown}(\mathcal{F}^d_2 + \mathcal{F}_3),
\end{align}
where $\{\mathcal{F}^d_1, \mathcal{F}^d_2, \mathcal{F}^d_3\}$ represent the intermediate modulated features that encapsulate reciprocal task cues. Finally, the aggregated outputs at each level are passed to two task-specific prediction heads, yielding the high-fidelity restored image $\hat{I}''$ and the precise saliency mask $\hat{M}''$.

\subsection{Loss Function}
%During training, UniV2D takes the degraded image $\hat{I}$, the corresponding clear image $\hat{I}_\textit{gt}$, and the reference saliency mask $M_\textit{gt}$ for supervision. In the inference phase, the model only requires the degraded image $\hat{I}$ as input.
To achieve collaborative optimization of both underwater image restoration and salient object detection, we employ a multi-task training objective that supervises the network at both the initialization and refinement stages.

\textbf{Saliency Detection Loss.}
To supervise the final saliency map $\hat{M}''$, we employ a composite loss consisting of Binary Cross-Entropy (BCE) and Intersection-over-Union (IoU) terms. While BCE ensures pixel-level classification accuracy, the IoU loss encourages global structural completeness and compensates for the imbalanced distribution of salient pixels. The refined saliency loss $\mathcal{L}_{\textit{mask}}^{\textit{fin}}$ is defined as follows:
\begin{equation}\label{Mask_Loss_function}
\mathcal{L}^{\textit{fin}}_{\textit{mask}}= \mathcal{L}_{\textit{BCE}}(\hat{M}^{''}, {M}_\textit{gt})+\mathcal{L}_{\textit{IoU}}(\hat{M}^{''}, {M}_\textit{gt}).
\end{equation}

\textbf{Structure-Aware Reconstruction Loss.}
To ensure pixel-wise accuracy and structural fidelity of the restored image $\hat{I}''$, we define the reconstruction loss $\mathcal{L}_{\textit{content}}$ by combining the $\ell_1$ norm with the Structural Similarity Index (SSIM). This combination balances local pixel intensity consistency with global perceptual structure:
\begin{equation}\label{Content_loss_function}
\mathcal{L}_{\textit{content}}=\left\|\hat{I}^{''}-{I}_\textit{gt}\right\|_{1}+(1-SSIM(\hat{I}^{''},{I}_\textit{gt})).
\end{equation}

\textbf{Perceptual Loss.}
We incorporate a VGG-based perceptual loss to minimize the feature-level discrepancy between the restored image $\hat{I}''$ and the ground truth $I_{\textit{gt}}$. By leveraging a pre-trained VGG-16 network \cite{Simonyan2014Very}, we ensure that the restored results align with high-level human visual perception. The perceptual loss $\mathcal{L}_{\textit{vgg}}$ is expressed as:
\begin{equation}
\mathcal{L}_{\textit{vgg}} = \sum_{k \in \{1,2,3\}} \left\| \Phi^{(k)}_{\textit{vgg}}(\hat{I}^{''}) - \Phi^{(k)}_{\text{vgg}}({I}_\textit{gt}) \right\|_1,
\end{equation}
where $\Phi^{(k)}{\textit{vgg}}(\cdot)$ denotes the feature activations from the $k$-th selected layer of the VGG-16 backbone.

\textbf{Total Training Objective.}
The final training objective integrates the content reconstruction, perceptual similarity, and saliency prediction losses from both the initialization and refinement stages to facilitate end-to-end multi-task learning:
\begin{equation}
\mathcal{L}_{\textit{total}} = \alpha \left( \mathcal{L}_{\textit{mask}}^{\textit{pre}} + \mathcal{L}_{\textit{mask}}^{\textit{fin}} \right) + \mathcal{L}_{\textit{content}} + \mathcal{L}_{\textit{vgg}},
\end{equation}
where $\mathcal{L}_{\textit{mask}}^{\textit{pre}}$ is the initial saliency loss from the SCSM module. The hyperparameters are empirically set to $\alpha=0.5$ to balance the convergence rates and significance of the two tasks.

%At inference time, given an underwater input image $J$, the model produces the final restored image $\hat{I}^{''}$ and the refined saliency map $\hat{M}^{''}$ through a two-stage process consisting of hierarchical initialization and saliency-guided refinement.
\begin{figure*}[!htp]
	\setlength{\abovecaptionskip}{0.05cm}
	\setlength{\belowcaptionskip}{0.0cm}
	\centering
	\includegraphics[width=1.0\textwidth]{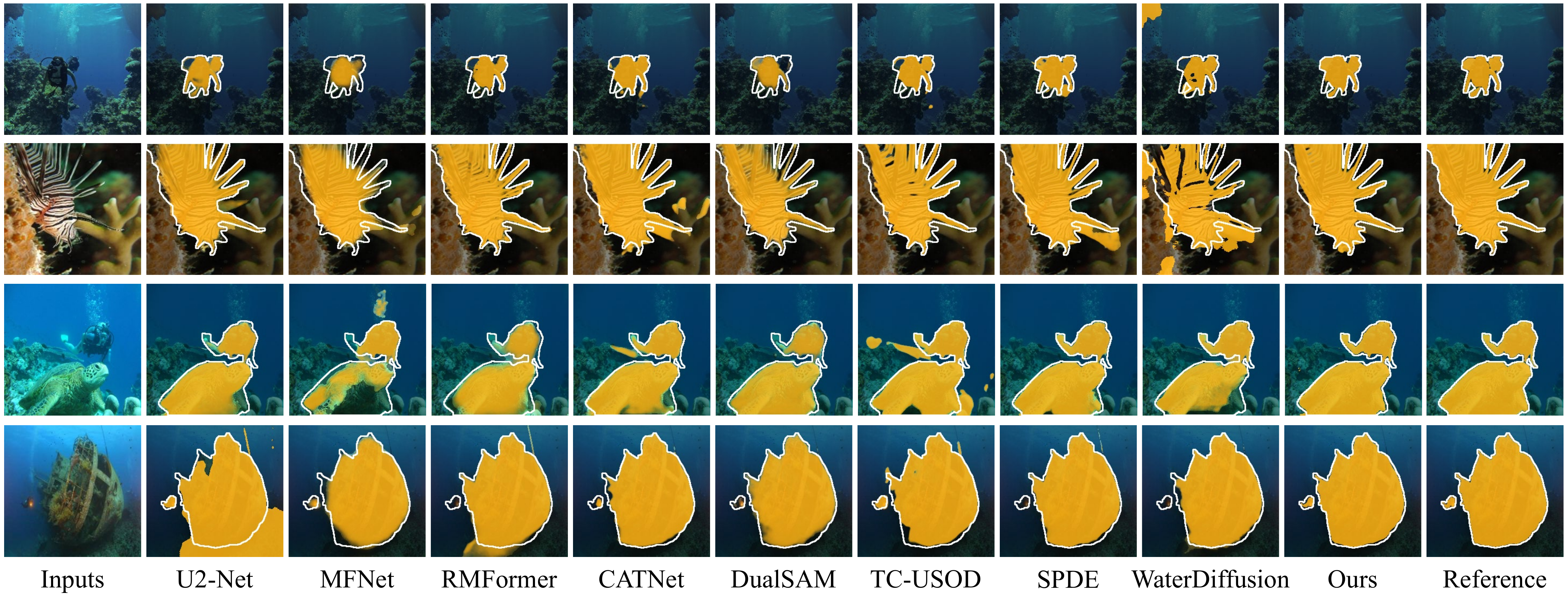}
	\caption	
	{Qualitative comparisons between UniV2D and SOTA methods across diverse biological categories and object scales (\textit{e.g.}, diver, lionfish, turtle, and shipwreck). Orange regions represent predicted salient masks, while the white contours indicate ground-truth annotations.}
	\label{4-2Qualitative_Evaluation_of_Saliency_Detection}
    \vspace{-0.3cm}
\end{figure*}
\section{Experiments}\label{Experiments}
\subsection{Experimental Setups}
\textbf{Implementation Details.} 
We implement the proposed UniV2D using the PyTorch framework and conduct training on two NVIDIA GeForce RTX 4090 GPUs for $150$ epochs. The batch size and patch size are set to 32 and $256 \times 256$, respectively. We employ the Adam optimizer with an initial learning rate of $1 \times 10^{-4}$. Detailed hyperparameter configurations and network specifications are provided in the supplementary material to ensure reproducibility.

\textbf{Benchmark Datasets.}
Our experiments evaluate the collaborative tasks of underwater image enhancement (UIE) and salient object detection (USOD) using several representative benchmarks. For the UIE task, we utilize UIEB \cite{Li2020UIEB}, LSUI \cite{Peng2023U-shapeLSUI}, and UWscene \cite{Islam2020EUVP} datasets; following standard protocols, UIEB and LSUI are split into training and testing sets, with 790 and 3,779 images for training, and 100 (Test-U100) and 500 (Test-L500) images reserved for evaluation, respectively. For the USOD task, we employ USOD10K \cite{Hong2025USOD10k}, USOD \cite{Islam2021SVAM}, and SUIM \cite{Islam2020SUIM} benchmarks. All datasets consist of real-world underwater images paired with high-quality reference maps, and for those with predefined splits (\textit{e.g.}, UWscene and SUIM), we strictly adhere to their original configurations. To ensure a fair comparison, all baseline methods are retrained on these same datasets using their official settings to maintain benchmarking consistency.

\textbf{Evaluation Metrics.}
For the UIE task, we adopt five widely-used metrics to assess restoration quality. These include three full-reference metrics: Structural Similarity (SSIM), Peak Signal-to-Noise Ratio (PSNR), and Perceptual Color Quality Index (PCQI), along with two no-reference metrics: UIQM \cite{Panetta2016UIQM} and UCIQE \cite{Yang2015UCIQE}. For the USOD task, we evaluate performance using four standard metrics: S-measure ($S_{\alpha}$) \cite{Fan2017S-measure}, weighted F-measure ($F_{\beta}^{w}$) \cite{Achanta2009F-measure}, max E-measure ($E_{\phi}^{m}$) \cite{Fan2018E-measure}, and Mean Absolute Error ($M_{AE}$) \cite{Perazzi2012MAE}. These metrics collectively provide a comprehensive evaluation of both low-level visual fidelity and high-level semantic localization.

\subsection{Comparison with State-of-the-Arts}
We compare the proposed UniV2D model with eight state-of-the-art UIE methods, including MLLE \cite{Zhang2022MLLE}, U-Shape \cite{Peng2023U-shapeLSUI}, Sem-UIR \cite{Huang2023Semi-UIR}, WWPF \cite{Zhang2024WWPF}, TCTL-Net \cite{Li2023TCTL_Net}, SMDR \cite{Zhang2024SMDR}, UWIRE \cite{Chang2025UWIRE}, and WaterDiffusion \cite{Chang2025Waterdiffusion}. For salient object detection evaluation, we compare UniV2D with eight well-known USOD methods, namely U2-Net \cite{Qin2020U2Net}, MFNet \cite{Piao2021Mfnet}, RMFormer \cite{Deng2023RMFormer}, CATNet \cite{Sun2023Catnet}, DualSAM \cite{Zhang2024DualSAM}, TC-USOD \cite{Hong2025USOD10k}, SPDE \cite{Jin2025SPDE}, and WaterDiffusion \cite{Chang2025Waterdiffusion}. Notably, WaterDiffusion is the only one capable of simultaneously performing both UIE and USOD tasks within a unified framework, making it our primary baseline for multi-task evaluation.

\begin{table*}[!htp]\small
    \setlength{\abovecaptionskip}{0.1cm}
	\setlength{\belowcaptionskip}{0.0cm}
	\renewcommand\arraystretch{1.0}
	\tabcolsep 0.01in
	\caption{Quantitative evaluation of each UIE method on three underwater datasets: Test-U100, Test-L500, and UWscene. The best and second-best results are highlighted in \textbf{bold} and \underline{underline}, respectively.}
	\centering
	\begin{tabular}{lccccc|ccccc|ccccc}
		\hline
		\multirow{2}{*}{Method} & \multicolumn{5}{c|}{\textbf{Test-U100}} & \multicolumn{5}{c|}{\textbf{Test-L500}} & \multicolumn{5}{c}{\textbf{UWscene}} \\
		\cline{2-16}
		& SSIM$\uparrow$ & PSNR$\uparrow$ & PCQI$\uparrow$ & UIQM$\uparrow$ & UCIQE$\uparrow$ & SSIM$\uparrow$ & PSNR$\uparrow$ & PCQI$\uparrow$ & UIQM$\uparrow$ & UCIQE$\uparrow$ & SSIM$\uparrow$ & PSNR$\uparrow$ & PCQI$\uparrow$ & UIQM$\uparrow$ & UCIQE$\uparrow$ \\
		\hline
		MLLE   & 0.631 & 16.66 & 0.449 & 4.216 & 0.595 & 0.632 & 19.20 & 0.413 & 4.402 & 0.602 & 0.592 & 14.98 & 0.415 & 4.403 & 0.613 \\
		U-shape & 0.816 & 21.83 & 0.836 & 4.760 & 0.563 & 0.849 & 24.63 & 0.808 & 4.885 & 0.564 & 0.820 & 22.89 & 0.776 & 4.997 & 0.579 \\
		Semi-UIR & 0.852 & 23.37 & 0.713 & 4.851 & 0.605 & 0.815 & 24.07 & 0.588 & 4.889 & 0.606 & 0.831 & 21.02 & 0.651 & 4.981 & 0.623 \\
		WWPF  & 0.711 & 17.58 & 0.511 & 4.504 & 0.603 & 0.680 & 19.42 & 0.434 & 4.634 & 0.609 & 0.648 & 15.80 & 0.428 & 4.641 & 0.627 \\
		TCTL-Net &\textbf{0.906} &\underline{{25.36}} & 0.817 & 4.747 & 0.602 & 0.838 & 22.83 & 0.713 & 4.857 & 0.595 & 0.855 & 23.54 & 0.718 & 4.940 & 0.606 \\
		SMDR  & 0.873 & 23.58 & 0.807 & 4.801 & 0.594 & \underline{{0.875}} & \underline{{25.22}} & 0.734 & 4.854 & 0.575 &\underline{{0.887}} & 23.84 & 0.672 & 5.001 & 0.613 \\
		UWIRE   & 0.751 & 23.65 & 0.678 & 4.850 & \underline{{0.616}} & 0.716 & 19.73 & 0.587 & \underline{{4.948}} & \textbf{{0.623}} & 0.655 & 17.54 & 0.579 & 4.970 & \textbf{{0.638}} \\
		Initialization  &0.860 &23.76 &\underline{{0.848}} &4.733 &0.597 &0.872 &24.99 &\underline{{0.817}} &4.856 &0.576 &0.861 &\underline{{25.03}} &\underline{{0.790}} &\underline{5.006} &0.572 \\
		UniV2D & \underline{{0.896}} & \textbf{{25.85}} & \textbf{{0.852}} & \textbf{{4.913}} & \textbf{{0.619}} & \textbf{{0.888}} & \textbf{{27.31}} & \textbf{{0.828}} & \textbf{{4.957}} & \underline{{0.611}} & \textbf{{0.894}} & \textbf{{26.34}} & \textbf{{0.818}} & \textbf{{5.067}} & \underline{{0.629}} \\
		\hline
	\end{tabular}
	\label{Visual_Enhancement_Metrics}
    \vspace{-0.1cm}
\end{table*}
\begin{table*}[!htp]\small
	\setlength{\abovecaptionskip}{0.1cm}
	\setlength{\belowcaptionskip}{0.0cm}
	\renewcommand\arraystretch{1.0}
	\tabcolsep 0.09 in
	\centering
	\caption{Quantitative evaluation of each USOD method on three underwater datasets: USOD10K, USOD, and SUIM. The best and second-best results are highlighted with \textbf{bold} and \underline{underline}, respectively.}
	\begin{tabular}{lcccc|cccc|cccc}
		\hline
		\multirow{2}{*}{Method} & \multicolumn{4}{c|}{\textbf{USOD10K}} & \multicolumn{4}{c|}{\textbf{USOD}} & \multicolumn{4}{c}{\textbf{SUIM}} \\
		\cline{2-13}
		&$S_{\alpha}\uparrow$ &$F_{\beta}^{w}\uparrow$ &$E_{\phi}^{m}\uparrow$ &$M_{AE}\downarrow$
		&$S_{\alpha}\uparrow$ &$F_{\beta}^{w}\uparrow$ &$E_{\phi}^{m}\uparrow$ &$M_{AE}\downarrow$
		&$S_{\alpha}\uparrow$ &$F_{\beta}^{w}\uparrow$ &$E_{\phi}^{m}\uparrow$ &$M_{AE}\downarrow$ \\
		\hline
		\text{SUIM-Net} &0.797 & 0.783 & 0.856 & 0.0821 & 0.769 & 0.717 & 0.826 & 0.1159 &0.777 & 0.601 & 0.804 & 0.0957 \\
		\text{U2-Net} &0.895 & 0.844 & 0.937 & 0.0351 & 0.888 &0.862 & 0.923 & 0.0541 & 0.816 & 0.718 & 0.899 & 0.0781 \\
		\text{MFNet}  &0.843 & 0.731 & 0.915 & 0.0513 & 0.842 & 0.776 & 0.911 & 0.0758 & 0.781 & 0.642 & 0.820 & 0.0974 \\
		\text{RMFormer}  & 0.867 & 0.828 & 0.910 & 0.0439 & 0.880 & 0.876 & 0.912 & 0.0542 & 0.808 & 0.713 & 0.870 & 0.0835 \\
		\text{CATNet}  & 0.890 & 0.862 & 0.949 & 0.0299 & 0.878 & 0.865 & 0.925 & 0.0522 & 0.831 & 0.768 & 0.895 & 0.0618 \\
		\text{DualSAM}  &0.916 & 0.909 & 0.959 & 0.0218 & \underline{{0.906}} & 0.908 & \underline{{0.940}} & 0.0397 & 0.853 & 0.804 & 0.909 & 0.0446 \\
		\text{TC-USOD}  & 0.912 & 0.905 & 0.953 & 0.0236 & 0.906 & \underline{{0.896}} & 0.937 & \underline{{0.0376}} & {0.867} & 0.833 & \underline{{0.934}} & 0.0409 \\
		\text{SPDE}  & \underline{{0.923}} & \underline{{0.912}} & \underline{{0.961}} & \textbf{{0.0203}} & 0.900 & 0.888 & 0.934 & 0.0438 & \underline{{0.876}} & \textbf{{0.863}} & 0.926 & \underline{{0.0382}} \\
		Initialization &0.894 &0.867 &0.947 &0.0379 &0.888 &0.877 &0.927	&0.0507 &0.840 &0.754 &0.920 &0.0562\\
		UniV2D   & \textbf{{0.927}} & \textbf{{0.915}} & \textbf{{0.965}} & \underline{{0.0211}} & \textbf{{0.912}} & \textbf{{0.901}} & \textbf{{0.946}} & \textbf{{0.0371}} & \textbf{{0.882}} & \underline{{0.858}} & \textbf{{0.949}} & \textbf{{0.0351}} \\
		\hline
	\end{tabular}
	\label{Saliency_Detection_Metrics}
	\vspace{-0.1cm}
\end{table*}
\textbf{Qualitative evaluation.} Fig. \ref{4-1Qualitative_Evaluation_of_Visual_Enhancement} illustrates representative restoration results on various underwater scenes. While most approaches yield visually pleasing enhancements, the proposed UniV2D consistently achieves results that most closely align with the reference images in terms of color fidelity, content preservation, and overall contrast. Given the subjectivity of individual visual perception, we further incorporate two widely adopted quantitative image quality metrics (SSIM and PSNR) to objectively evaluate performance. The consistently higher values achieved by our UniV2D underscore its effectiveness in visual restoration tasks. Furthermore, Fig. \ref{4-2Qualitative_Evaluation_of_Saliency_Detection} presents salient object detection results across the USOD10K, USOD, and SUIM datasets. Compared to existing techniques, UniV2D demonstrates remarkable robustness in accurately localizing salient regions and preserving fine-grained boundaries, even in challenging environments with low contrast or cluttered backgrounds.

%While most methods produce visually plausible enhancements, they often suffer from residual color casts or over-exposure in complex lighting. In contrast, UniV2D consistently yields outcomes that most closely align with the reference images, exhibiting superior color fidelity, texture preservation, and balanced contrast. Given the subjectivity of individual visual perception, we further incorporate two widely adopted quantitative image quality metrics (SSIM and PSNR) to objectively evaluate performance. The consistently higher values achieved by our UniV2D underscore its effectiveness in visual restoration tasks. Furthermore, Fig. \ref{4-2Qualitative_Evaluation_of_Saliency_Detection} presents salient object detection results across the USOD10K, USOD, and SUIM datasets. Compared to existing techniques, UniV2D demonstrates remarkable robustness in accurately localizing salient regions and preserving fine-grained boundaries, even in challenging environments with low contrast or cluttered backgrounds.

\textbf{Quantitative evaluation.} 
The quantitative results for underwater image enhancement and salient object detection across multiple public datasets are presented in Table \ref{Visual_Enhancement_Metrics} and Table \ref{Saliency_Detection_Metrics}, respectively. Our proposed UniV2D consistently achieves the best or second-best performance across nearly all metrics, particularly attaining the highest average scores in key full-reference indices. This superior performance demonstrates UniV2D's robustness in complex underwater scenes, outperforming both task-specific SOTA models and the dual-task WaterDiffusion. These gains stem from our hierarchical dual-stage design: the SCSM and MACR modules establish robust structural priors during initialization, while the refinement stage utilizes CLFM for deep semantic alignment. This synergy facilitates mutual reinforcement, where enhanced visual features improve saliency localization, which in turn provides sharper structural guidance for final high-fidelity restoration.

%Such gains are primarily attributed to our hierarchical dual-branch architecture, where the initialization stage leverages SCSM and MACR modules to establish initial structural priors, while the Task-Reciprocal Refinement stage utilizes CLFM modules for deep semantic alignment. This collaborative design facilitates effective mutual reinforcement, ensuring that clearer visual features lead to more precise saliency localization, which in turn provides sharper structural guidance for final high-fidelity restoration.

\subsection{Evaluation of Model Efficiency}
\begin{figure}[!tp]
	\setlength{\abovecaptionskip}{0.1cm}
	\setlength{\belowcaptionskip}{0.0cm}
	\centering
	\includegraphics[width=0.480\textwidth]{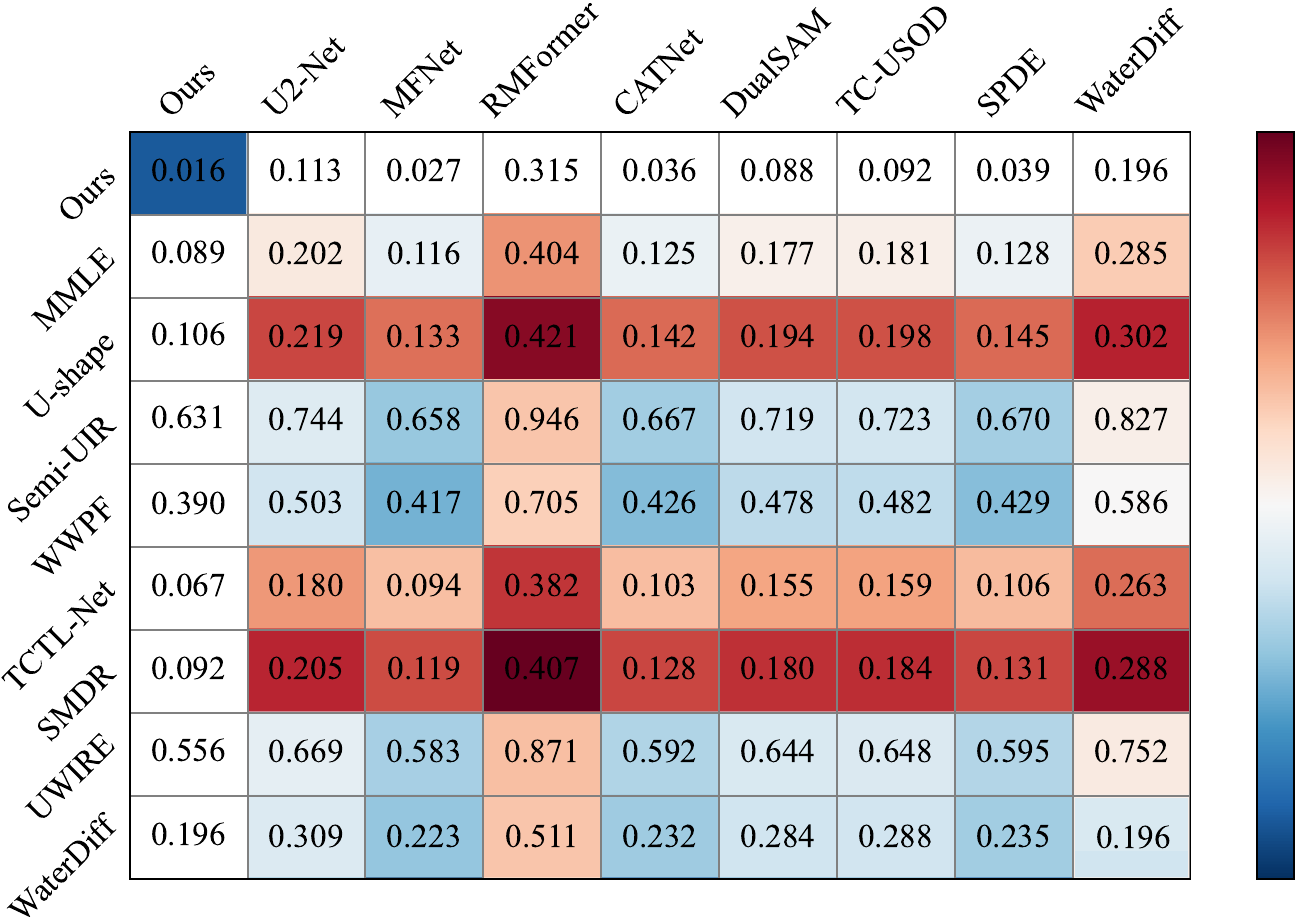}
	\caption
	{Efficiency evaluation of each compared method in terms of inference time. The center region indicates the summed time of two tasks at the corresponding position.}
	\label{4-4Elapsed_time_of_each_method}
	\vspace{-0.2cm}
\end{figure}
\textbf{Inference Efficiency.} 
We first evaluate the efficiency of UniV2D by comparing its inference time with UIE and USOD methods. As illustrated in Fig. \ref{4-4Elapsed_time_of_each_method}, we present the inference times for individual UIE and USOD tasks along the axes, while the center region represents the cumulative time of their linear cascades (\textit{i.e.}, UIE followed by USOD). Our UniV2D achieves a remarkably low inference time of 0.016s per image, which is significantly faster than even the most efficient cascaded combination (e.g., TCTL-Net and MFNet at 0.094s). This nearly six-fold speedup validates that our unified architecture effectively eliminates the redundant computational overhead inherent in conventional multi-stage pipelines.
\begin{table}[!tp]\small
	\setlength{\abovecaptionskip}{0.1cm}
	\setlength{\belowcaptionskip}{0.1cm}
	\renewcommand\arraystretch{1.0}
	\tabcolsep 0.022 in
	\caption{Efficiency evaluation of each compared method in terms of Parameters (M) and FLOPs (G).}
	\label{FLOPs_and_Params}
	\centering
	\begin{tabular}{lcc|lcc}
		\hline
		\multicolumn{3}{c|}{\textbf{UIE Methods}} &\multicolumn{3}{c}{\textbf{USOD Methods}}\\
		\hline
		Config. &Params. &FLOPs &Config. &Params. &FLOPs \\
		\hline
		MLLE &\textit{Null} &\textit{Null} &U2-Net &44.01 &37.65 \\ 
		U-Shape &65.60 	&66.20  &MFNet &36.36 &38.69	\\	
		Semi-UIR  &12.78 &36.46 &RMFormer &174.19 &563.14\\
		WWPF  &\textit{Null} &\textit{Null} &CATNet &110.40 &67.80 \\
		TCTL-Net &99.72 &56.51 &DualSAM &159.95 &325.70  \\
		SMDR  &12.26  &46.59 &TC-USOD  &117.64 	&29.64 \\
		UWIRE  &\textit{Null} &\textit{Null} &SPDE &115.66 &41.46\\
		WaterDiffusion  &55.49  &169.56  &WaterDiffusion  &55.49  &169.56 \\
		Ours  &\textbf{13.29} &\textbf{31.23} &Ours &\textbf{13.29} &\textbf{31.23}	\\
		%Ours  &21.52 &37.23 &Ours &21.52 &37.23	\\
		\hline
	\end{tabular}
	\vspace{-0.2cm}
\end{table}

\textbf{Parameters and FLOPs.} To further evaluate model complexity, we conduct a comparative analysis of parameters (M) and FLOPs (G) against deep learning-based competitors, as summarized in Table \ref{FLOPs_and_Params}. Despite its dual-task capability, UniV2D maintains a highly compact footprint with only 13.29M parameters and 31.23G FLOPs. Notably, compared to the multi-task WaterDiffusion, UniV2D achieves a 76\% reduction in parameters and an 81\% decrease in computational overhead. Furthermore, UniV2D demonstrates superior efficiency even when compared to many single-task models.

%\begin{figure}[!htp]
	%\setlength{\abovecaptionskip}{0.1cm}
	%\setlength{\belowcaptionskip}{-0.2cm}
	%\centering
	%\includegraphics[width=0.40\textwidth]{Figs/4-4Elapsed_time_of_each_method.pdf}
	%\caption
	%{Efficiency evaluation of each compared method in terms of inference time. The center region indicates the summed time of two tasks at the corresponding position.}
	%\label{4-4Elapsed_time_of_each_method}
	%\vspace{-0.1cm}
%\end{figure}

\subsection{Ablation Study}
To verify the contribution of each core component in UniV2D, we conduct comprehensive ablation experiments focusing on the initialization and refinement stages.

\textbf{Ablation Study of MACR Module.}
Table \ref{Ablation_study_of_MACR} presents the impact of the Mask-Aware Content Restoration (MACR) module across four saliency metrics. The inclusion of MACR leads to consistent improvements in $S_{\alpha}$, $F_{\beta}^{w}$, and $E_{\phi}^{m}$, alongside a significant reduction in $M_{AE}$. These gains demonstrate that integrating mask-aware features effectively suppresses background interference and enhances high-level semantic localization, thereby improving saliency prediction by leveraging refined visual cues.
\begin{table}[!htp]\small
	\setlength{\abovecaptionskip}{0.1cm}
	\setlength{\belowcaptionskip}{0.1cm}
	\renewcommand\arraystretch{1.0}
	\tabcolsep 0.065 in
	\centering
	\caption{Quantitative ablation of the MACR module.}
	\begin{tabular}{cc|cccc}
		\hline
		-w/o MACR &-w/ MACR  & $S_{\alpha}\uparrow$ &$F_{\beta}^{w}\uparrow$ &$E_{\phi}^{m}\uparrow$ & $M_{AE}$ $\downarrow$\\
		\hline
		$\checkmark$ & &0.888 &0.859 &0.941 &0.0421 \\
		&$\checkmark$  &\textbf{0.925} &\textbf{0.910} &\textbf{0.963} &\textbf{0.0217} \\
		\hline
	\end{tabular}
	\label{Ablation_study_of_MACR}
	\vspace{-0.1cm}
\end{table}

\textbf{Ablation Study of SCSM Module.} 
Table \ref{Ablation_study_of_SCSM} summarizes the performance contribution of the self-calibrated saliency masking (SCSM) module to the visual restoration task. The introduction of SCSM yields superior SSIM, PSNR, UIQM, and UCIQE scores, validating its effectiveness in optimizing perceived contrast and color constancy. These results indicate that by providing fine-grained saliency-guided attention, the module enables the network to adaptively concentrate on informative regions, resulting in higher restoration fidelity and overall visual quality.
\begin{table}[!htp]\small
	\setlength{\abovecaptionskip}{0.1cm}
	\setlength{\belowcaptionskip}{0.1cm}
	\renewcommand\arraystretch{1.0}
	\tabcolsep 0.046 in
	\centering
	\caption{Quantitative ablation of the SCSM module.}
	\begin{tabular}{cc|cccc}
		\hline
		-w/o SCSM &-w/ SCSM  & SSIM$\uparrow$ & PSNR$\uparrow$ & UIQM$\uparrow$ & UCIQE$\uparrow$\\
		\hline
		$\checkmark$ & &0.863 &24.979 &4.969 &0.574 \\
		&$\checkmark$ &\textbf{0.893}  &\textbf{26.492} &\textbf{5.031} &\textbf{0.623} \\
		\hline
	\end{tabular}
	\label{Ablation_study_of_SCSM}
	\vspace{-0.1cm}
\end{table}

\begin{figure}[htp]
	\setlength{\abovecaptionskip}{0.1cm}
	\setlength{\belowcaptionskip}{0.1cm}
	\centering
	\includegraphics[width=0.480\textwidth]{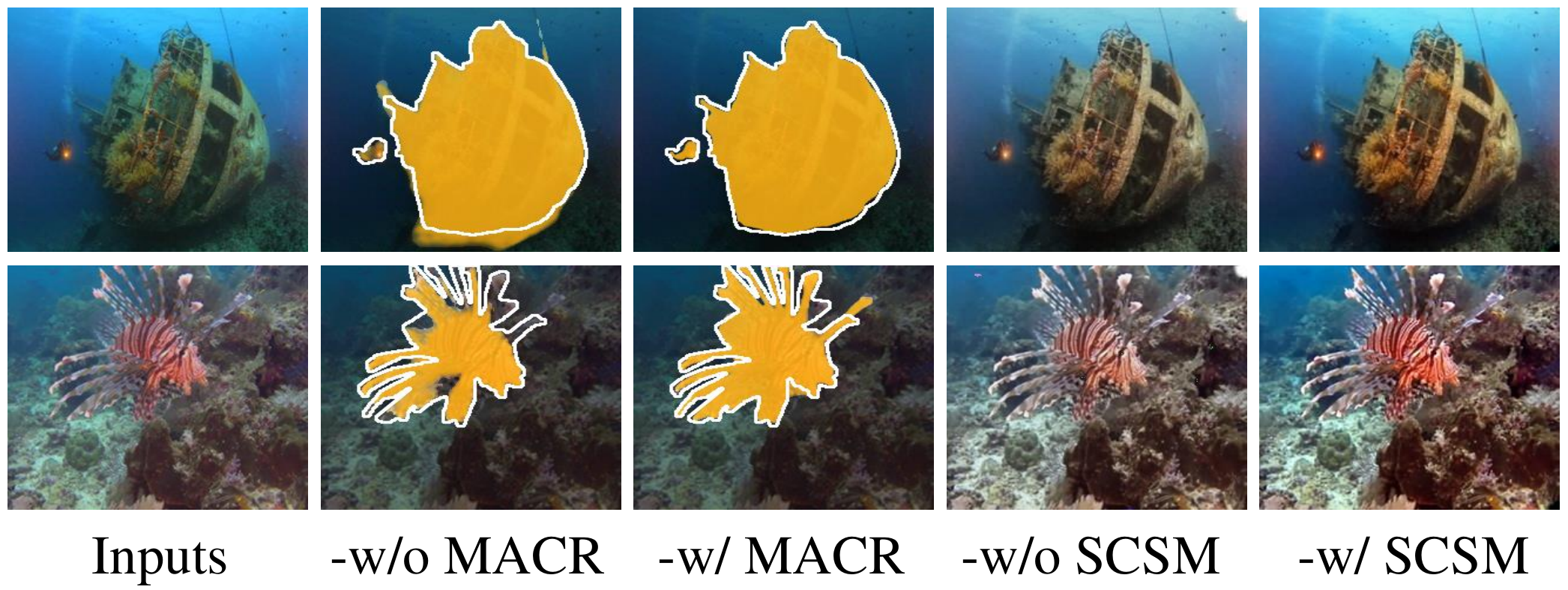}
	\caption
	{Visual ablation of the SCSM and MACR modules.}
	\label{4-5Visual_Abalation_Mask_Image}
	\vspace{-0.1cm}
\end{figure}

\textbf{Visual ablation of MACR and SCSM.} 
Fig. \ref{4-5Visual_Abalation_Mask_Image} provides a qualitative comparison to illustrate the structural contributions of both modules. Specifically, the MACR module enhances the regional completeness and boundary sharpness of the predicted saliency masks. The SCSM module effectively mitigates underwater color casts and improves structural clarity in the restored images. These visual outcomes further confirm the effectiveness of the dual-branch interaction in fostering task-specific enhancements.

\textbf{Effectiveness of Saliency Mask Flow.}
Table \ref{Ablation_study_of_SMF} evaluates the effectiveness of the Saliency Mask Flow (SMF) during both the initialization and refinement stages. In both phases, the incorporation of SMF consistently optimizes all saliency metrics, yielding higher $S_{\alpha}$, $F_{\beta}^{w}$, and $E_{\phi}^{m}$ scores alongside a diminished $M_{AE}$. Notably, the more significant performance leap in the refinement stage suggests that SMF effectively propagates semantic cues across stages, thereby enforcing saliency consistency and contributing to more robust and accurate object localization.

\begin{table}[htp]\small
	\setlength{\abovecaptionskip}{0.1cm}
	\setlength{\belowcaptionskip}{0.1cm}
	\renewcommand\arraystretch{1.1}
	\tabcolsep 0.076 in
	\centering
	\caption{Effectiveness of Saliency Mask Flow (SMF) across initialization and refinement stages.}
	\begin{tabular}{c|c|cccc}
		\hline
		& Config. & $S_{\alpha}\uparrow$ &$F_{\beta}^{w}\uparrow$ &$E_{\phi}^{m}\uparrow$ & $M_{AE}\downarrow$ \\
		\hline
		\multirow{2}{*}{Initialization} 
		& -w/o SMF & 0.879 & 0.838 & 0.928 & 0.0365 \\
		& -w/ SMF 
		& \textbf{0.894} 
		& \textbf{0.867} 
		& \textbf{0.947} 
		& \textbf{0.0319} \\
		\hline
		\multirow{2}{*}{Refinement} 
		& -w/o SMF & 0.885 & 0.853 & 0.939 & 0.0340 \\
		& -w/ SMF 
		& \textbf{0.925} 
		& \textbf{0.910} 
		& \textbf{0.963} 
		& \textbf{0.0217} \\
		\hline
	\end{tabular}
	\label{Ablation_study_of_SMF}
	\vspace{-0.1cm}
\end{table}

\section{Conclusion}\label{Conclusion}
This paper presents UniV2D, a novel unified framework designed for the joint optimization of USOD and UIE. Diverging from conventional cascaded or multi-stage baselines, UniV2D integrates both tasks within a single, cohesive network, enabling low-level enhancement and high-level perception to mutually reinforce each other. Specifically, the proposed hierarchical dual-branch initialization utilizes the SCSM module to estimate robust saliency priors, while the MACR module restores visual content under the explicit guidance of these predicted masks. Subsequently, a task-reciprocal refinement stage employs the CLFM module to enforce deep semantic alignment and preserve fine-grained structural details across multiple scales. Extensive experiments on several benchmarks demonstrate that UniV2D consistently achieves state-of-the-art performance in both UIE and USOD tasks. %Moreover, our model exhibits superior efficiency and compactness, providing a practical and effective solution for real-time underwater visual perception.

\small
\bibliographystyle{ieeenat_fullname}
\bibliography{Ref}

%\appendix

\end{document}